\documentclass[lettersize,journal]{IEEEtran}
\usepackage{amsmath,amssymb,amsfonts}
\usepackage{algorithm}
\usepackage{algpseudocode}
\usepackage{array}
\usepackage[caption=false,font=normalsize,labelfont=sf,textfont=sf]{subfig}
\usepackage{textcomp}
\usepackage{stfloats}
\usepackage{url}
\usepackage[breaklinks,colorlinks,citecolor=citeblue]{hyperref}
\usepackage{graphicx}
\usepackage{cite}
\usepackage{booktabs}
\usepackage{multirow}
\usepackage[table,xcdraw,svgnames,dvipsnames]{xcolor}
\definecolor{tvcgblue}{rgb}{0.94, 0.78, 1.0}
\definecolor{citeblue}{rgb}{0.21,0.49,0.74}
\newcommand{\pub}[1]{{\tiny\textcolor{red}{#1}}}

\begin{document}

\title{MMOE: Modernizing Diffusion Transformers \\with Efficient Expert Design}

\author{Yanhao Jia, Jiepeng Wang, Haibin Huang, Chi Zhang, \\ Erik Cambria, \textit{Fellow, IEEE}, Xuelong Li, \textit{Fellow, IEEE}
\thanks{Yanhao Jia and Jiepeng Wang contributed equally to this work. This work was done during Yanhao's internship at TeleAI. Corresponding authors: Erik Cambria and Xuelong Li.}
\thanks{Yanhao Jia and Erik Cambria are with the College of Computing and Data Science, Nanyang Technological University, Singapore, 639798 (E-mail: yanhao002@e.ntu.edu.sg, cambria@ntu.edu.sg).}
\thanks{Jiepeng Wang, Haibin Huang, Chi Zhang and Xuelong Li are with the Institute of Artificial Intelligence of China Telecom (TeleAI), Shanghai, China, 200232 (E-mail: jiepeng@connect.hku.hk, huanghb28@chinatelecom.cn, zhangc120@chinatelecom.cn, xuelongli@ieee.org).}}

\markboth{IEEE Transactions on Visualization and Computer Graphics,~Vol.~XX, No.~XX, Month~Year}%
{Author \MakeLowercase{\textit{et al.}}: MMOE: Modernizing Diffusion Transformers with Efficient Expert Design}

\IEEEtitleabstractindextext{%
\begin{abstract}
Modern large language models scale successfully by pairing capacity growth with efficiency, keeping per-token and deployment costs under control as capacity grows. AIGC Foundation Models (AFMs), especially diffusion-transformer backbones, have begun to adopt sparse experts, but recent efforts mostly enlarge total parameter counts and sparsity ratios without importing the efficiency mechanisms that made LLM scaling practical, so generation quality is seldom balanced against training and deployment cost. This raises a natural question: can the architectural principles behind efficient LLM scaling be adapted to AFMs in a more balanced way? We introduce ModernMOE (MMOE), a modernization of SiT-style diffusion transformers that systematically adapts routed experts, shared and lightweight experts, gate-residual routing, and attention-residual information reuse to AIGC generation. Rather than treating MoE as a single plug-in replacement, MMOE studies how different modern expert components affect convergence, efficiency, and generation quality when composed inside a diffusion transformer. Every experiment in this paper is trained on a single eight-GPU H100 node with batch size 256 for 400k steps, an accessible single-machine budget. Under matched training and sampling protocols and at this budget, MMOE reaches lower FID at every recorded checkpoint, that is, it converges faster per training step, than dense and intermediate sparse-expert baselines, and among the sparse variants it attains the best quality-cost balance. Routing analysis further shows stable expert specialization across depth, substantial use of lightweight routes, and modest step-to-step routing changes during denoising. These results suggest that AFMs can follow the balanced scaling path of LLMs by importing proven efficiency designs, rather than by simply increasing total parameters and sparsity ratios.
\end{abstract}

}

\maketitle
\IEEEdisplaynontitleabstractindextext
\IEEEpeerreviewmaketitle

\section{Introduction}
\label{sec:introduction}

\begin{figure*}[t]
    \centering
    \includegraphics[width=\linewidth]{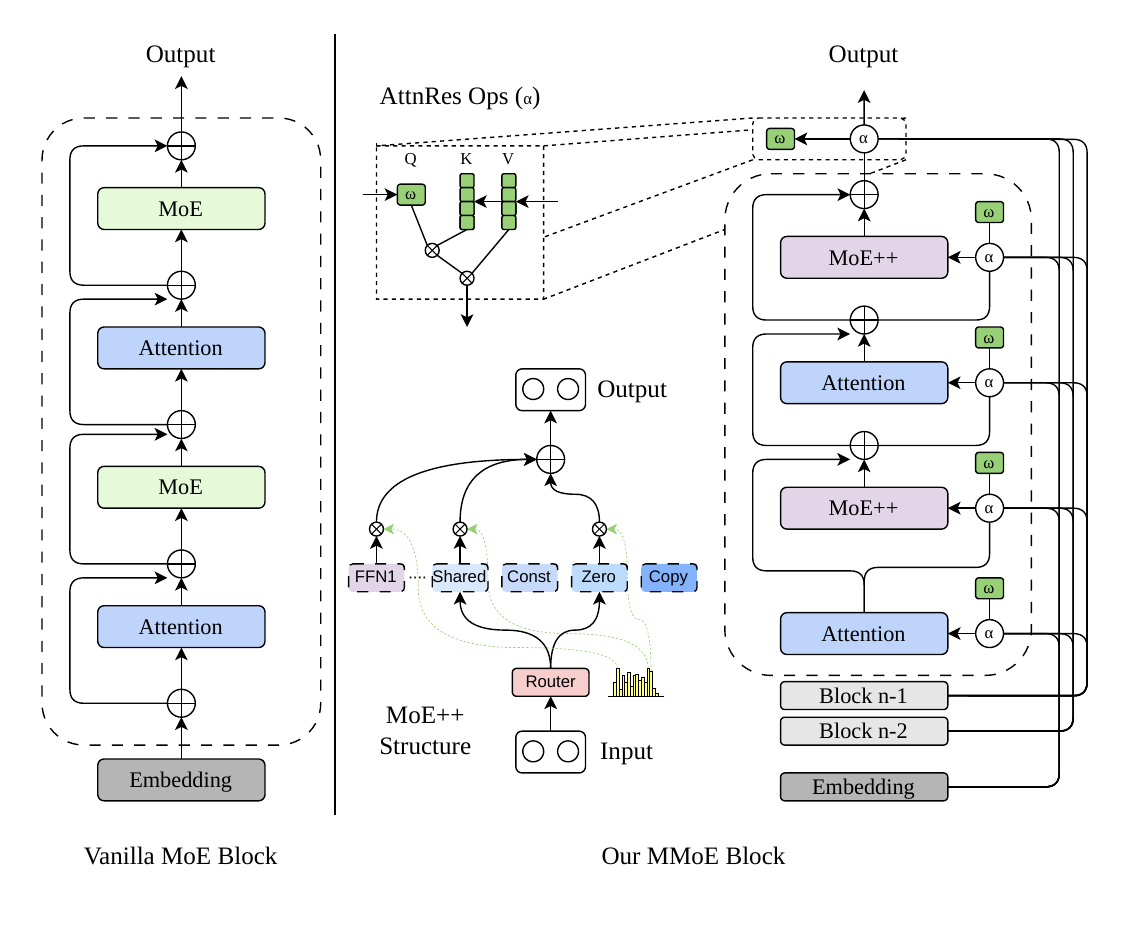}
    \vspace{-15mm}
    \caption{Architecture of MMOE. Left: a vanilla MoE block that routes each token to a small pool of feed-forward experts. Right: the MMOE block, which adds MoE++ lightweight experts (copy, zero, and constant), gate-residual routing, and attention-residual aggregation over previously completed block states before both the attention and the expert sub-layer.}
    \label{fig:mmoe}
\end{figure*}

\IEEEPARstart{T}{ransformer} scaling has reshaped both language modeling and visual generation, but the two fields have followed different architectural trajectories. Modern large language models increasingly rely on sparse and high-throughput computation: sparsely gated experts decouple total capacity from activated computation~\cite{shazeer2017outrageously}, large-scale systems such as GShard and Switch Transformers make conditional computation practical at scale~\cite{lepikhin2021gshard,fedus2022switch}, and recent LLMs further popularize top-$k$ expert routing as a standard way to enlarge model capacity while controlling per-token cost~\cite{jiang2024mixtral}. More recent MoE designs go beyond simply adding routed FFN experts. For example, MoE++ introduces zero-computation experts such as zero, copy, and constant experts, together with gate-residual routing, to improve both efficiency and effectiveness~\cite{jin2025moepp}. These developments suggest that the progress of foundation models is no longer only a matter of increasing parameter count, but also a matter of designing how computation is activated, skipped, reused, and routed.

AIGC Foundation Models (AFMs), especially diffusion-transformer backbones, have also benefited from transformer scaling. DiT demonstrates that replacing convolutional denoisers with transformers yields a scalable generative backbone~\cite{peebles2023scalable}, while SiT provides a closely related scalable interpolant-transformer formulation for flow and diffusion generation~\cite{ma2024sit}. However, scaling AFMs remains expensive because generation requires repeated denoising, and each denoising step processes many visual tokens through deep transformer blocks. Recent work has therefore begun to explore MoE for diffusion transformers. DiT-MoE studies shared expert routing and expert-level balancing for sparse diffusion transformers~\cite{fei2024scaling}. EC-DIT uses expert-choice routing to allocate computation according to heterogeneous generation complexity~\cite{sun2025ecdit}. Diff-MoE introduces time-aware and space-adaptive experts for denoising stages and spatial tokens~\cite{cheng2025diffmoe}. Other studies investigate routing competition, explicit routing guidance, and practical recipes for training diffusion MoE models~\cite{yuan2025race,wei2025routing,liu2025efficientmoe}. Together, these works show that MoE is a promising direction for AFMs. A recurring pattern, however, is to scale mainly by enlarging the total expert pool and the sparsity ratio: DiT-MoE trains a 4.1B-parameter model for up to 7M iterations at batch size 1024~\cite{fei2024scaling}, and Race-DiT scales to 2.79B parameters over 1.7M iterations~\cite{yuan2025race}. These efforts inherit the bigger-pool mindset while largely leaving aside the efficiency mechanisms, such as zero-computation experts, gate-residual routing, and explicit control of activated computation, that let LLM MoE trade capacity against per-token and deployment cost. This leaves an unresolved design question: which of these LLM efficiency designs actually transfer to diffusion-transformer generation, and how should they be combined when training cost and deployment efficiency matter?

This paper takes a modernization perspective. ConvNeXt revisited a classical vision backbone and showed that carefully importing modern design choices can close much of the gap to a newer architectural paradigm~\cite{liu2022convnet}. We ask an analogous question for AFMs: can a standard SiT-style diffusion transformer be modernized with recent LLM MoE designs to improve generation quality without increasing the effective computation and training burden? This question is practically important because simply increasing total parameters and sparsity ratios can make AFMs difficult to train, route, communicate, and deploy on resource-constrained devices. Instead of treating MoE as a single replacement for dense FFNs, we study a sequence of recent expert designs, including routed experts, shared experts, lightweight zero-computation experts, gate-residual routing, and attention-residual information reuse. We test these components under a controlled diffusion-transformer setting and integrate the effective parts into a unified architecture.

We propose ModernMOE (MMOE), a modernized sparse-expert diffusion transformer for AFMs, illustrated in Figure~\ref{fig:mmoe}. MMOE keeps the denoising objective, conditioning interface, latent representation, and sampler unchanged, and modifies only the internal computation of transformer blocks. Starting from a SiT-style backbone, we progressively introduce standard routed MoE, shared-expert computation, MoE++-style lightweight experts, and attention-residual aggregation. This progressive design allows us to separate the effect of each component rather than attributing all gains to a monolithic architecture. The resulting model allocates expensive MLP computation selectively, provides cheap expert routes when a heavy update is unnecessary, and reuses cross-depth information without adding a new generative objective or sampling procedure.

Every experiment in this paper is trained on a single eight-GPU H100 node with batch size 256 for 400k steps, an accessible single-machine budget. At this budget, MMOE reaches lower FID at every recorded checkpoint, that is, it converges faster per training step, than the dense and intermediate sparse-expert baselines, and among the sparse variants it attains the best quality-cost balance; a direct empirical comparison against prior diffusion-MoE methods, which are trained at far larger batch sizes and iteration counts, is left as ongoing work. Ablation studies further indicate that each modern LLM-inspired MoE design contributes to performance, convergence, or speed when adapted to AFMs. Beyond aggregate generation quality, we analyze routing behavior during denoising and observe strong block-wise specialization, depth-dependent lightweight-expert use, and low adjacent-step routing switch rates. These observations support the view that sparse expert modernization can serve not only as a way to increase nominal capacity, but also as a mechanism for stable adaptive computation allocation in generative foundation models.

Our contributions are summarized as follows:
\begin{itemize}
    \item We formulate AFM scaling as a modernization problem and systematically examine whether recent LLM MoE designs can improve diffusion-transformer generation without simply increasing total parameters, sparsity ratios, or training overhead.
    \item We propose ModernMOE, a SiT-style sparse-expert architecture that integrates routed experts, shared experts, lightweight zero-computation experts, gate-residual routing, and attention-residual aggregation into a unified diffusion-transformer backbone.
    \item We provide controlled comparisons, ablations, efficiency analysis, and routing analysis, all within a single eight-GPU H100 node, showing that the proposed modernization path improves AFM performance and convergence at an accessible budget while revealing how modern expert components affect convergence, speed, and specialization.
\end{itemize}

\section{Related Work}
\label{sec:related-work}

\subsection{Diffusion Models and Flow Matching}
\label{subsec:rw-diffusion-flow}

Diffusion models define generation as the reverse of a gradual noising process and have become a central paradigm for high-fidelity image synthesis~\cite{ho2020denoising}. Score-based generative models further formulate this process through stochastic differential equations, connecting denoising, score estimation, stochastic samplers, and probability-flow ODEs under a unified continuous-time view~\cite{song2021score}. A systematic study of the design space of these models further disentangles noise scheduling, network preconditioning, and sampling to markedly improve sample quality~\cite{karras2022elucidating}. Latent diffusion models reduce the cost of this iterative generation process by moving denoising from pixel space to a compressed autoencoder latent space, enabling high-resolution generation with a more tractable token budget~\cite{rombach2022high}.

Flow-based formulations provide a closely related perspective. Flow Matching trains continuous normalizing flows by directly regressing vector fields along probability paths, avoiding simulation of a reverse diffusion process during training~\cite{lipman2023flow}. Stochastic interpolants further unify flow and diffusion models by describing generative paths that bridge noise and data distributions through continuous-time interpolating processes~\cite{albergo2025stochastic}. Recent transformer-based generative backbones build on these ideas: U-ViT casts all inputs as tokens with a ViT backbone and long skip connections~\cite{bao2023all}, DiT replaces convolutional denoisers with transformer blocks over latent patches~\cite{peebles2023scalable}, and SiT studies scalable interpolant transformers for both flow- and diffusion-based generation~\cite{ma2024sit}; complementary training techniques such as aligning intermediate features with pretrained visual representations further ease diffusion-transformer training~\cite{yu2025representation}. These transformer backbones now reach well beyond class-conditional image synthesis: DiT-based systems unify multimodal image generation and understanding~\cite{wang2025mmgen}, extend to controllable and multi-shot video generation~\cite{xi2026omnivdiff,xi2025ctrlvdiff,wu2026edustory}, and support neural 3D geometry and appearance reconstruction~\cite{liu2023nero}. The broader push toward general multimodal foundation models also drives progress in cross-modal understanding and generation~\cite{jia2025uni, wu2025query, wu2026towards, jia2026seeing, cambria2026senticnet, jiatowards}. MMOE is orthogonal to the choice between diffusion and flow matching: it keeps the generative objective, conditioning interface, and sampler fixed, and instead modernizes the internal computation of the diffusion-transformer backbone.

\subsection{MoEs in LLMs and Diffusion Models}
\label{subsec:rw-moe}

Mixture-of-experts models increase total model capacity by activating only a subset of parameters for each token. The sparsely-gated MoE layer established this conditional-computation paradigm~\cite{shazeer2017outrageously}, and large-scale systems such as GShard and Switch Transformers made sparse expert routing practical for foundation-model training~\cite{lepikhin2021gshard,fedus2022switch}. GLaM shows that sparsely activated experts can match dense models at a fraction of the training cost~\cite{du2022glam}, while ST-MoE improves the training stability and transferability of sparse expert models~\cite{zoph2022stmoe}. Recent LLMs further demonstrate that top-$k$ routed FFN experts can provide a favorable capacity-throughput tradeoff~\cite{jiang2024mixtral}. Beyond standard routed FFNs, modern MoE designs increasingly explore more fine-grained efficiency mechanisms: DeepSeekMoE finely segments experts and isolates shared experts to reduce redundant computation~\cite{dai2024deepseekmoe}, expert-choice routing lets experts select tokens to balance load~\cite{zhou2022mixture}, and large-scale systems such as DeepSeek-V3 adopt auxiliary-loss-free load balancing for efficient training~\cite{deepseekai2024deepseekv3}. MoE++ introduces zero-computation experts, including zero, copy, and constant experts, so that not every selected route needs to execute a full MLP~\cite{jin2025moepp}, and mixture-of-depths applies a related activate-or-skip principle across transformer layers~\cite{raposo2024mixture}. Sparse experts have likewise been scaled to vision transformers~\cite{riquelme2021scaling}. Attention Residuals studies selective reuse of previous layer states, offering another way to improve information flow and memory behavior in deep transformer models~\cite{kimiteam2026attentionresiduals}.

MoE has also started to appear in diffusion-transformer research. Large sparse diffusion transformers have been explored as a scaling path for high-capacity generative models~\cite{fei2024scaling, kukreja2026forge}. EC-DIT uses expert-choice routing for adaptive computation in text-to-image diffusion transformers~\cite{sun2025ecdit}. Diff-MoE introduces time-aware and space-adaptive experts to match denoising stages and spatial token complexity~\cite{cheng2025diffmoe}, while DiffMoE (distinct from Diff-MoE above) proposes dynamic token selection with a batch-level global token pool and a capacity predictor~\cite{shi2025diffmoe}. Other works investigate routing competition, explicit routing guidance, and practical architecture choices for training diffusion MoE models~\cite{yuan2025race,wei2025routing,liu2025efficientmoe}. These studies show that sparse experts are promising for AIGC generation, but most of them focus on a particular routing strategy or scaling recipe. MMOE instead asks a complementary question: which modern MoE components developed in LLMs remain effective when transferred to diffusion transformers, and how can they be combined into a practical AFM backbone?

\subsection{Scaling Law with AIGC Foundation Models}
\label{subsec:rw-scaling}

Scaling has been a dominant driver of progress in AIGC Foundation Models. Latent diffusion reduces the spatial cost of image generation~\cite{rombach2022high}, and transformer denoisers such as DiT and SiT show that generative quality can improve with larger transformer backbones and longer training~\cite{peebles2023scalable,ma2024sit}. Sparse diffusion transformers push this trend further by increasing total parameter capacity while activating only part of the network for each token~\cite{fei2024scaling}. From a scaling-law perspective, this suggests that AFMs should be evaluated not only by nominal parameter count, but also by activated parameters, routing overhead, memory footprint, training throughput, and deployment efficiency.

The contrast with LLM scaling is instructive. In LLMs, sparse-expert scaling became practical because capacity growth was paired with mechanisms that hold per-token and deployment cost roughly constant: routing decouples activated computation from total capacity, while zero-computation experts and gate-residual routing let the model skip or reuse computation instead of always executing a full expert. Current AFM scaling instead tends to enlarge the total expert pool and the sparsity ratio while importing few of these efficiency mechanisms. The asymmetry is also one of budget: prior sparse AFMs are trained with billions of parameters for 1.7M to 7M iterations at batch size 1024~\cite{fei2024scaling,yuan2025race}, whereas the study in this paper fits on a single eight-GPU H100 node at batch size 256 for 400k steps. This matters because sampling repeatedly applies the backbone over many denoising steps, so routing overhead, expert dispatch and communication, load balancing, and deployment execution all recur at every step. Serving foundation models efficiently under edge and resource-constrained conditions is itself an active research direction~\cite{shao2025ai,an2026ai,chen2026generativetransmissionrethinkingcomputation}, which reinforces the need to control activated computation and communication rather than only total capacity. A useful AFM scaling path should therefore improve quality under realistic compute budgets rather than merely enlarge the total expert pool. MMOE follows this principle by modernizing the block-level computation: it allocates expensive MLP computation selectively, provides cheaper expert paths when possible, and reuses cross-depth information to improve the performance-cost tradeoff.

\section{Method}
\label{sec:method}

MMOE imports four efficiency mechanisms proven in LLM MoE, namely routed experts, shared and lightweight zero-computation experts, gate-residual routing, and attention-residual reuse, into the SiT backbone, changing only how block-level computation is activated and reused so that capacity can grow without a proportional increase in activated computation. Figure~\ref{fig:mmoe} contrasts the resulting MMOE block with a vanilla MoE block. MMOE is built on the SiT backbone and keeps the external SiT interface unchanged: the model receives latent tokens $x$, timestep $t$, and class label $y$, embeds the latent patches with a fixed positional embedding, forms the conditioning vector $c=e_t(t)+e_y(y)$, and returns the unpatchified denoising prediction. The difference from the dense SiT backbone is entirely inside the transformer stack. Each block is split into an attention sub-layer and a MoE++ MLP sub-layer, and both sub-layers can receive an attention-residual aggregation over previously completed block states. The full forward pass maintains three recurrent states: a list of completed block states, a current partial block state, and a gate residual passed across expert layers.

\subsection{MoE++ expert layer}
\label{subsec:mmoe-expert-layer}

The MLP sub-layer in MMOE is implemented by a MoE++ expert layer. Given hidden states $H \in \mathbb{R}^{B \times T \times D}$, the layer first flattens the batch and token dimensions into $N=BT$ token representations. A router produces expert logits for every token. The default router is a two-layer gate,
\begin{equation}
\begin{aligned}
    g(h) &= W_2 \tanh(W_1 h),
\end{aligned}
\end{equation}
where $h \in \mathbb{R}^D$ is a token feature, $W_1 \in \mathbb{R}^{8E \times D}$ and $W_2 \in \mathbb{R}^{E \times 8E}$ are learned gate weights, $D$ is the model width, and $E$ is the number of experts. If a previous gate residual is available, the router maps it through a learned linear transformation and adds it to the current logits:
\begin{equation}
\begin{aligned}
    \ell_i &= g(h_i) + A r_i^{\mathrm{prev}} .
\end{aligned}
\end{equation}
Here $A \in \mathbb{R}^{E \times E}$ maps the previous gate residual $r_i^{\mathrm{prev}} \in \mathbb{R}^{E}$ into the current logit space, and the resulting logits satisfy $\ell_i \in \mathbb{R}^{E}$. The updated logits $\ell_i$ are returned as the next gate residual. This implementation makes routing decisions depend not only on the current token state, but also on the routing tendency inherited from the previous expert layer.

The router selects top-$2$ experts for each token. Under the default non-Mixtral gating mode, probabilities are computed by softmax over all experts, the top-$2$ probabilities are selected and renormalized, and the zero expert receives zero gate weight when selected. The MoE output is accumulated by dispatching only the selected tokens to each expert:
\begin{equation}
\begin{aligned}
    \operatorname{MMOE}_{\mathrm{mlp}}(h_i)
    &=
    \sum_{e \in \mathcal{T}_2(i)}
    p_{i,e} f_e(h_i),
\end{aligned}
\end{equation}
where $\mathcal{T}_2(i)$ denotes the selected expert indices and $p_{i,e}$ denotes the post-renormalization routing weight.\footnote{When the zero expert is among the top-$2$ for a token, its weight is set to $0$ before renormalization, so the remaining weight is assigned entirely to the other selected expert and that token is effectively processed by a single expert.} In code, this is implemented with per-expert token indices, expert-specific forward calls, and an \texttt{index\_add} accumulation into the output tensor.\footnote{In the default routing mode the dispatched token feature is pre-scaled by the top-$k$ factor, so each expert is evaluated at $2 h_i$ rather than $h_i$ before its output is weighted by $p_{i,e}$.}

The expert pool contains both heavy and lightweight experts. If the number of experts is $E$, the implementation constructs $E-4$ standard MLP experts, two constant experts, one copy expert, and one zero expert. A heavy expert is a standard transformer MLP with hidden width determined by the MLP ratio. The copy and zero experts are
\begin{equation}
\begin{aligned}
    f_{\mathrm{copy}}(h)&=h,
    \qquad
    f_{\mathrm{zero}}(h)=0 .
\end{aligned}
\end{equation}
Each constant expert learns a vector $a_j$ and predicts an input-dependent mixture between the token feature and this constant vector:
\begin{equation}
\begin{aligned}
    f_{\mathrm{const},j}(h)
    &=
    \pi_{j,0}(h)h + \pi_{j,1}(h)a_j, \\
    \pi_j(h)
    &=
    \operatorname{softmax}(W_jh).
\end{aligned}
\end{equation}
Here $a_j \in \mathbb{R}^{D}$ is a learned constant vector and $W_j \in \mathbb{R}^{2 \times D}$ produces the two-way mixture weights $\pi_j(h) \in \mathbb{R}^2$. Thus, selected routes do not always trigger a full MLP. Some tokens can use cheap identity, zero, or constant transformations, which is the core implementation mechanism for reducing unnecessary heavy expert computation.

\subsection{MMOE transformer block}
\label{subsec:asmoe-block}

Each MMOE block follows the adaLN-Zero structure of SiT but separates attention and MLP computation into two functions. Given a block input $z$ and conditioning vector $c$, the block first produces six modulation vectors:
\begin{equation}
\begin{aligned}
    (&\Delta_{\mathrm{msa}}, S_{\mathrm{msa}}, G_{\mathrm{msa}}, \\
    &\quad \Delta_{\mathrm{mlp}}, S_{\mathrm{mlp}}, G_{\mathrm{mlp}})
    =
    \operatorname{MLP}_{\mathrm{adaLN}}(c).
\end{aligned}
\end{equation}
The attention sub-layer applies layer normalization, adaLN modulation, self-attention, and gated residual addition:
\begin{equation}
\begin{aligned}
    z_{\mathrm{attn}}
    &=
    z
    +
    G_{\mathrm{msa}} \odot
    \operatorname{Attn}\Big( \\
    &\quad
    \operatorname{mod}
    \big(
        \operatorname{LN}(z),
        \Delta_{\mathrm{msa}},
        S_{\mathrm{msa}}
    \big)
    \Big).
\end{aligned}
\end{equation}
The block caches $(\Delta_{\mathrm{mlp}},S_{\mathrm{mlp}},G_{\mathrm{mlp}})$ so that the MLP sub-layer does not recompute the adaLN projection. The MLP sub-layer then applies the MoE++ expert layer to the modulated token features and updates the gate residual:
\begin{equation}
\begin{aligned}
    m, r^{\mathrm{next}}
    &=
    \operatorname{MMOE}_{\mathrm{mlp}}\Big( \\
    &\quad
    \operatorname{mod}
    \big(
        \operatorname{LN}(z_{\mathrm{mlp}}),
        \Delta_{\mathrm{mlp}},
        S_{\mathrm{mlp}}
    \big), r^{\mathrm{prev}}
    \Big),
\end{aligned}
\end{equation}
\begin{equation}
\begin{aligned}
    z_{\mathrm{out}} &= z_{\mathrm{mlp}} + G_{\mathrm{mlp}} \odot m .
\end{aligned}
\end{equation}
Here $z_{\mathrm{mlp}}$ is the input to the MLP sub-layer, which may be either the post-attention state or an attention-residual aggregation of previous block states, as described next.

\subsection{Attention-residual forward process}
\label{subsec:asmoe-forward}

MMOE maintains a list of completed block states $\mathcal{C}$ and a current partial block state $u$. Before each attention sub-layer and before each MLP sub-layer, the model optionally aggregates information from $\mathcal{C}$ and $u$. If no completed state exists, the current partial state is used directly. Otherwise, the implementation stacks the completed block states $s_1,\ldots,s_M \in \mathcal{C}$ and the current partial state $u$,
\begin{equation}
\begin{aligned}
    V &= [s_1,\ldots,s_M,u],
\end{aligned}
\end{equation}
normalizes them with an RMSNorm, and applies a learned pseudo-query $q$ along the state dimension:
\begin{equation}
\begin{aligned}
    \alpha_i
    &=
    \operatorname{softmax}_i
    \Big(
        q^\top \operatorname{RMSNorm}(V_i)
    \Big), \\
    \operatorname{AttnRes}(V)
    &=
    \sum_i \alpha_i V_i .
\end{aligned}
\end{equation}
The model uses separate projection and RMSNorm parameters for attention-residual aggregation before attention and before MLP at every layer. This gives the block two opportunities to reuse previous completed states: one before self-attention and another before expert computation.

The complete forward process is summarized in Algorithm~\ref{alg:mmoe-forward}. After every block group, the current partial state is appended to the completed-state list and the partial state is reset to zeros.\footnote{Immediately after a boundary the partial state is reset to zeros but is still included as the last entry of $V$ in the next aggregation, so the first sub-layer of each new block group attends over the completed states together with a zero placeholder.} In the implementation, the boundary interval is \texttt{block\_size / 2}. This grouped update avoids attending to every intermediate layer state while still allowing later blocks to reuse representations from earlier completed groups. The implementation keeps the latest block output as the final transformer representation, so the reset operation at a block boundary does not erase the output used by the final prediction head. After the final transformer block, MMOE applies the standard SiT final layer and unpatchifies the result. If sigma prediction is enabled, the output channels are split and only the denoising prediction is returned. No additional denoising objective, conditioning branch, or sampler is introduced.

\begin{algorithm}[t]
\caption{MMOE forward process}
\label{alg:mmoe-forward}
\begin{algorithmic}[1]
\Require latent input $x$, timestep $t$, label $y$
\State $u \gets \operatorname{PatchEmbed}(x)+\operatorname{PosEmbed}$
\State $c \gets e_t(t)+e_y(y)$
\State $\mathcal{C}\gets [\ ]$, $r\gets \mathrm{None}$, $h\gets u$
\For{block $i=1,\ldots,L$}
    \State $z_{\mathrm{attn}}\gets u$ if $\mathcal{C}$ is empty,
    \Statex \hspace{\algorithmicindent} else $\operatorname{AttnRes}_{i}^{\mathrm{attn}}(\mathcal{C},u)$
    \State $(u,\mathrm{cache})\gets \operatorname{AttentionSubLayer}_i(z_{\mathrm{attn}},c)$
    \State $z_{\mathrm{mlp}}\gets u$ if $\mathcal{C}$ is empty,
    \Statex \hspace{\algorithmicindent} else $\operatorname{AttnRes}_{i}^{\mathrm{mlp}}(\mathcal{C},u)$
    \State $(u,r)\gets \operatorname{MMOESubLayer}_i($
    \Statex \hspace{\algorithmicindent} $z_{\mathrm{mlp}},c,r,\mathrm{cache})$
    \State $h\gets u$
    \If{$i$ reaches a block-group boundary}
        \State append $u$ to $\mathcal{C}$
        \State reset $u$ to zeros
    \EndIf
\EndFor
\State \Return $\operatorname{Unpatchify}(\operatorname{FinalLayer}(h,c))$
\end{algorithmic}
\end{algorithm}

\section{Experiments}
\label{sec:experiments}

\subsection{Experiment Settings}
\label{subsec:setup}

We evaluate MMOE on class-conditional ImageNet-256 generation~\cite{deng2009imagenet}. All models are trained in the latent space of a Stable-Diffusion-style VAE~\cite{rombach2022high}. Generated images are decoded with the same VAE and evaluated with FID~\cite{heusel2017gans} using pytorch-fid; lower FID is better in all tables. Unless otherwise specified, we report EMA checkpoints, ODE sampling, 50 denoising steps, 50k generated samples, and classifier-free guidance~\cite{ho2022classifierfree}. The purpose of this setup is to isolate architectural changes: dataset preprocessing, latent representation, denoising objective, sampler, and evaluation metric are kept fixed across variants.

\begin{figure}[!t]
    \centering
    \includegraphics[width=\linewidth]{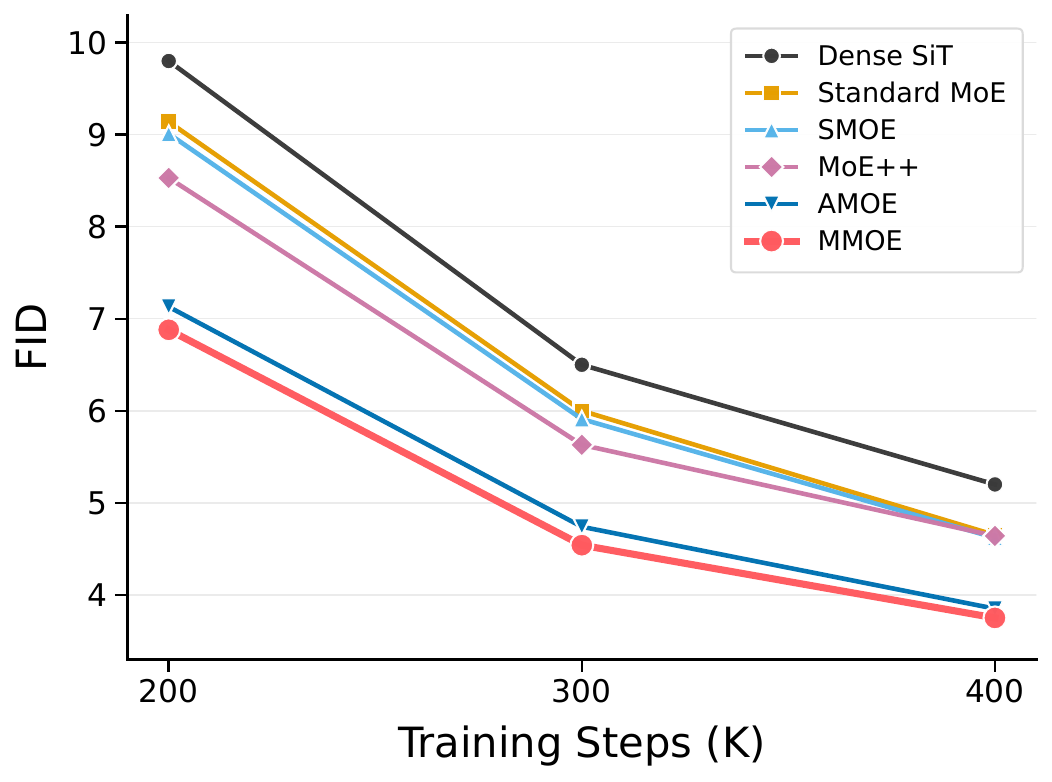}
    \caption{FID convergence on ImageNet-256 for XL/2 models along the modernization path from dense SiT to MMOE. MMOE reaches the lowest FID at every checkpoint, indicating faster convergence per training step. Lower is better.}
    \label{fig:fid-convergence}
\end{figure}

All variants use the same SiT training interface. The VAE downsampling factor maps ImageNet-256 images to $32 \times 32$ latent grids with four channels, and the ImageNet-512 runs in Table~\ref{tab:imagenet512} map $512 \times 512$ images to $64 \times 64$ latents through the same pipeline. Training uses the linear interpolant, velocity prediction, uniform timestep sampling, and class-label dropout with probability $0.1$ for classifier-free guidance. Every run in this paper is trained on a single eight-GPU H100 node with batch size 256, which keeps the entire modernization study within an accessible single-machine budget. Unless otherwise noted, models are trained for 400k optimization steps with AdamW, learning rate $1\times10^{-4}$, betas $(0.9,0.999)$, no weight decay, fp16 mixed precision, maximum gradient norm $1.0$, and EMA decay $0.9999$. For MoE variants that expose a load-balancing auxiliary loss, the loss is added with coefficient $0.01$. The evaluation script samples labels uniformly, uses the same VAE decoder for all models, and computes FID against precomputed ImageNet reference statistics. The main modernization comparison in Table~\ref{tab:main-imagenet256} uses 50-step ODE sampling at guidance scale 1.5 with 50k generated samples; the model-scale study in Table~\ref{tab:model-size} and the guidance study in Table~\ref{tab:cfg} use 250-step ODE sampling; and the sample-count and sampling-step sweeps in Table~\ref{tab:sample-count} and Table~\ref{tab:sampling-steps} vary those two axes as indicated.

We compare a controlled sequence of architectures. Each step introduces one design family from modern sparse expert models while keeping the surrounding SiT training recipe unchanged:
\begin{itemize}
    \item Dense SiT: the baseline with dense MLP blocks.
    \item Standard MoE: all MLP blocks are replaced by routed MLP experts, with a top-$k$ router and load-balancing auxiliary loss.
    \item Shared MoE~(SMOE): one shared dense expert is always active, and additional routed experts provide token-dependent specialization.
    \item MoE++: routed MLP experts are combined with copy, zero, and constant experts, together with gate-residual routing.
    \item AttnRes-MoE~(AMOE): standard routed MLP experts are combined with attention-residual aggregation.
    \item MMOE: MoE++-style lightweight experts are combined with attention-residual aggregation; this is the final modernized AIGC backbone described in Section~\ref{sec:method}.
\end{itemize}

\subsection{Comparison Results}
\label{subsec:main-results}

\begin{figure}[tb]
    \centering
    \includegraphics[width=\linewidth]{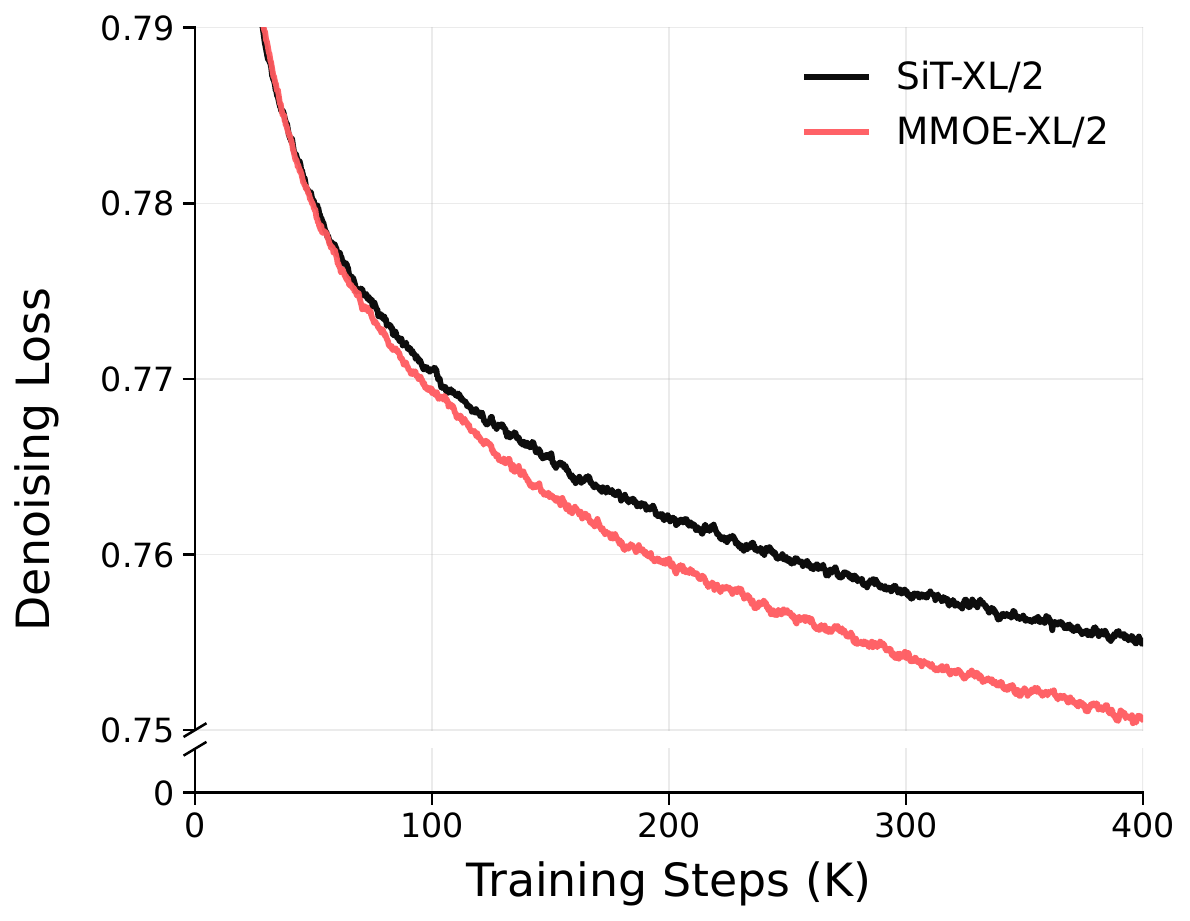}
    \caption{Denoising-loss convergence of dense SiT-XL/2 and MMOE-XL/2 on ImageNet-256 (0--400k steps). The broken y-axis zooms into the converged range: MMOE tracks the dense baseline early in training and then settles at a consistently lower loss, corroborating its faster FID convergence.}
    \label{fig:loss-convergence}
\end{figure}

\begin{table*}[t]
    \centering
    \caption{Modernization path on ImageNet-256 class-conditional generation for XL/2 models, from dense SiT to MMOE. FID uses EMA checkpoints, 50-step ODE sampling, guidance scale 1.5, and 50k samples; the last row reports wall-clock training time on a single eight-GPU H100 node. Lower FID is better.}
    \label{tab:main-imagenet256}
    \renewcommand{\arraystretch}{1.05}  
    \begin{tabular}{l|ccccc>{\columncolor{tvcgblue!60}}c}
        \toprule
        Training steps & SiT & Standard MoE & SMOE & MoE++ & AMOE & MMOE \\
        \midrule
        200k & 9.80 & 9.14 & 9.01 & 8.53 & 7.13 & 6.88 \\
        300k & 6.50 & 6.00 & 5.91 & 5.63 & 4.74 & 4.54 \\
        400k & 5.20 & 4.65 & 4.62 & 4.64 & 3.85 & 3.75 \\
        \midrule
        Training time & 23h & 54h & 45h & 46h & 120h & 67h \\
        \bottomrule
    \end{tabular}
\end{table*}

Table~\ref{tab:main-imagenet256} and Figure~\ref{fig:fid-convergence} show the main modernization path, in which dense SiT is progressively modernized with routed experts, shared-expert computation, lightweight experts, and attention-residual expert aggregation; the training-time row reports wall-clock time on a single eight-GPU H100 node. Dense SiT provides the reference point. Replacing dense MLPs with standard routed experts improves FID at all recorded checkpoints, indicating that sparse expert capacity is useful even without changing the diffusion objective or sampler. SMOE further stabilizes the expert architecture by preserving an always-active common computation path. MoE++ introduces lightweight copy, zero, and constant experts, which help at the early and middle training stages by allowing the router to select cheaper transformations for some tokens. Attention-residual aggregation gives the largest single jump among the measured components, suggesting that cross-depth information reuse is particularly beneficial for sparse expert diffusion transformers.

The final MMOE variant combines lightweight expert routes with attention-residual aggregation. It obtains the best FID throughout the recorded training trajectory while requiring less wall-clock training time than the AMOE variant in our measured setting. This supports the central hypothesis of the paper: the best performance-cost tradeoff does not come from sparse routing alone, but from composing routing with cheap expert paths and representation reuse. Figure~\ref{fig:loss-convergence} corroborates this from the optimization side: MMOE-XL/2 tracks dense SiT-XL/2 early in training and then reaches a consistently lower denoising loss for most of the 0--400k trajectory, so the FID gains are accompanied by faster convergence at the same single-node budget.

\begin{table*}[t]
    \centering
    \caption{MMOE alongside dense and sparse-expert diffusion transformers on class-conditional ImageNet-256. The SiT-XL/2 and MMOE-XL/2 rows are our own single-node runs; the others are quoted from the cited papers and are not directly comparable (see text).}
    \label{tab:external-baselines}
    \renewcommand{\arraystretch}{1.05}
    \begin{tabular}{c|cccc}
        \toprule
        Method & Total params & Activated params & Train iters & FID \\
        \midrule
        SiT-XL/2 (ours)~\cite{ma2024sit}                    & 676M   & -     & 400k    & 4.91 \\
        DyDiT-XL/2~\pub{ICLR2025}~\cite{zhao2025dynamic}    & 678M   & -     & 7M + Finetune    & 2.07 \\
        JiT-L/16~\pub{CVPR2026}~\cite{li2026back}           & 953M   & -     & 1M      & 2.29 \\
        DiT-MoE-XL/2~\pub{arXiv2024}~\cite{fei2024scaling}  & 4.1B   & 1.5B  & 7M      & 1.72 \\
        DiffMoE-L-E8~\pub{arXiv2025}~\cite{shi2025diffmoe}  & 1.18B  & 458M  & 7M      & 2.13 \\
        Race-DiT-XL/2~\pub{ICML2025}~\cite{yuan2025race}    & 2.79B  & 710M  & 1.7M    & 2.06 \\
        DSMoE-3B-E16~\pub{arXiv2025}~\cite{liu2025efficientmoe} & 2.96B  & 965M  & 1M      & 2.39 \\
        ProMoE-XL/2~\pub{ICLR2026}~\cite{wei2025routing}    & 1.57B  & 675M  & 500k    & 4.11 \\
        \midrule
        \rowcolor{tvcgblue!60} MMOE-XL/2 (ours)             & 1.57B  & $\le$770M  & \textbf{400k}    & 3.60 \\
        \bottomrule
    \end{tabular}
\end{table*}

The comparisons in Table~\ref{tab:main-imagenet256} are internal: they isolate the contribution of each modern expert component while holding the SiT training recipe fixed. To place MMOE in the broader landscape, Table~\ref{tab:external-baselines} places it alongside dense baselines and prior diffusion-MoE methods as reported in their papers, all using classifier-free guidance; the SiT-XL/2 and MMOE-XL/2 rows are our own single-node runs, while the others are quoted from the cited sources. These figures are not directly comparable to ours. The baselines are trained far longer and at much larger batch sizes: DiT-MoE and Race-DiT use batch size 1024 for 1.7M to 7M iterations, whereas our two rows use batch size 256 for 400k iterations on a single eight-GPU H100 node, roughly an order of magnitude less training. They also use their own samplers and sample counts, with DiT-MoE quoting FID-50k under 250-step DDPM sampling, DiffMoE-L-E8 FID-50k under 250-step flow sampling, and Race-DiT not stating its sampler; our two rows use 250-step ODE sampling at guidance scale 1.5. EC-DIT~\cite{sun2025ecdit} is omitted because it reports only text-to-image MS-COCO FID rather than class-conditional ImageNet. Under these as-reported protocols, several methods trained far longer reach lower FID, including DiT-MoE-XL/2 (1.72) and DiffMoE-L-E8 (2.13) after up to 7M iterations, Race-DiT-XL/2 (2.06), DyDiT-XL/2 (2.07), DSMoE-3B-E16 (2.39), and the dense JiT-L/16 (2.29), all far beyond our 400k-iteration budget. At a comparable 1.57B total-parameter budget, however, MMOE (3.60) already improves on ProMoE-XL/2 (4.11) despite fewer training iterations. We therefore do not claim state-of-the-art FID; the contribution of MMOE is the controlled modernization study and its quality-cost balance at a much smaller, single-machine training budget. A matched-protocol comparison that trains these baselines and MMOE to the same budget is left as ongoing work, and the present empirical claims are scoped to the dense and intermediate sparse-expert baselines in Table~\ref{tab:main-imagenet256}.

\begin{table}[tb]
    \centering
    \caption{Model-size comparison of dense SiT and MMOE at the S/2, B/2, and L/2 scales (250-step ODE, guidance scale 1.0, 400k steps). MMOE improves FID over the corresponding dense SiT model at every scale, so the modernization is not specific to the XL setting.}
    \label{tab:model-size}
    \renewcommand{\arraystretch}{1.05}
    \begin{tabular}{c|c>{\columncolor{tvcgblue!60}}c}
        \toprule
        Scale & SiT & MMOE \\
        \midrule
        S/2 & 60.0 & 51.0 \\
        B/2 & 37.2 & 26.9 \\
        L/2 & 22.3 & 14.7 \\
        \bottomrule
    \end{tabular}
\end{table}

Table~\ref{tab:model-size} evaluates whether the modernization remains useful below the XL setting. Under the 250-step, CFG-1.0 protocol, MMOE improves over the corresponding dense SiT model at S, B, and L scales. This suggests that the benefit of the proposed sparse expert modernization is not restricted to a single model width or depth.

\begin{table}[t]
    \centering
    \caption{Training convergence of a one-billion-parameter MoE++ model at the L/2 scale, with eight experts (four MLP plus two constant, one copy, and one zero) and top-2 routing. FID keeps decreasing throughout training, and for this checkpoint 250-step sampling further improves the 50-step result.}
    \label{tab:moepp-l2}
    \renewcommand{\arraystretch}{1.05}
    \begin{tabular}{c|cccc}
        \toprule
        Training steps & 200k & 300k & 400k & 400k (250 steps) \\
        \midrule
        FID & 11.69 & 7.74 & 5.81 & 5.47 \\
        \bottomrule
    \end{tabular}
\end{table}

Table~\ref{tab:moepp-l2} reports the training convergence of a one-billion-parameter MoE++ model at the L/2 scale, using eight experts (four standard MLP experts plus two constant, one copy, and one zero expert) with top-2 routing and the default 50-step evaluation protocol. The lightweight-expert design keeps reducing FID throughout training at this larger total-parameter budget, reaching 5.81 at 400k steps. For this checkpoint, increasing the number of ODE steps from 50 to 250 further lowers FID to 5.47, in contrast to the MMOE-XL/2 checkpoint in Table~\ref{tab:sampling-steps}, where additional steps did not help. This indicates that sampling-step sensitivity is checkpoint dependent and should be tuned per model rather than assumed constant.

We further check two robustness aspects of the XL/2 comparison, both at 400k steps under the 50-step ODE, guidance-scale-1.5, 50k-sample protocol. Table~\ref{tab:seed-robustness} reports seed robustness over three random seeds (0, 1, 2): dense SiT, SMOE, and MMOE reach FID $5.16\pm0.04$, $4.62\pm0.06$, and $3.72\pm0.05$, so the gaps between variants are far larger than the seed-induced standard deviation. These three-seed means agree, within that variance, with the single-run (seed-0) values reported elsewhere in the paper, such as the MMOE and SiT entries in Table~\ref{tab:main-imagenet256} (3.72 versus 3.75 and 5.16 versus 5.20), so the single-run rankings are stable. Table~\ref{tab:imagenet512} extends the dense-versus-MMOE comparison to ImageNet-512: MMOE again improves over dense SiT, reaching 11.72 versus 12.72 at 200k steps and 6.03 versus 6.58 at 400k steps.

\begin{table}[tb]
    \centering
    \caption{Seed-robustness of the XL/2 modernization path: FID mean $\pm$ standard deviation over three random seeds (0, 1, 2) at 400k steps. The between-variant gaps are much larger than the seed-induced deviation, so the single-run rankings are stable.}
    \label{tab:seed-robustness}
    \renewcommand{\arraystretch}{1.05}
    \begin{tabular}{l|cc>{\columncolor{tvcgblue!60}}c}
        \toprule
         & SiT & SMOE & MMOE \\
        \midrule
        FID (mean $\pm$ std) & 5.16$\pm$0.04 & 4.62$\pm$0.06 & 3.72$\pm$0.05 \\
        \bottomrule
    \end{tabular}
\end{table}

\begin{table}[tb]
    \centering
    \caption{ImageNet-512 comparison between dense SiT and MMOE, using the same VAE pipeline that maps $512\times512$ images to $64\times64$ latents. MMOE improves over dense SiT at both the 200k and 400k checkpoints.}
    \label{tab:imagenet512}
    \renewcommand{\arraystretch}{1.05}
    \begin{tabular}{c|cc}
        \toprule
        Model & FID (200k) & FID (400k) \\
        \midrule
        SiT                             & 12.72 & 6.58 \\
        \rowcolor{tvcgblue!60}MMOE      & 11.72 & 6.03 \\
        \bottomrule
    \end{tabular}
\end{table}

\subsection{Ablation Study}
\label{subsec:efficiency}

The modernization sequence in Table~\ref{tab:main-imagenet256} also serves as the component ablation. Standard MoE isolates sparse expert capacity, SMOE tests the benefit of an always-active common path, MoE++ evaluates lightweight copy/zero/constant experts and gate-residual routing, and AMOE isolates attention-residual aggregation. Each component contributes to the final performance-cost balance, although the progression is not strictly monotonic at every checkpoint: MoE++ helps mainly at the early and middle checkpoints and is essentially tied with SMOE at 400k (4.64 versus 4.62), whereas attention-residual aggregation and lightweight expert routing give the largest and most consistent gains and are especially important for MMOE.

\begin{figure}[tb]
    \centering
    \includegraphics[width=\linewidth]{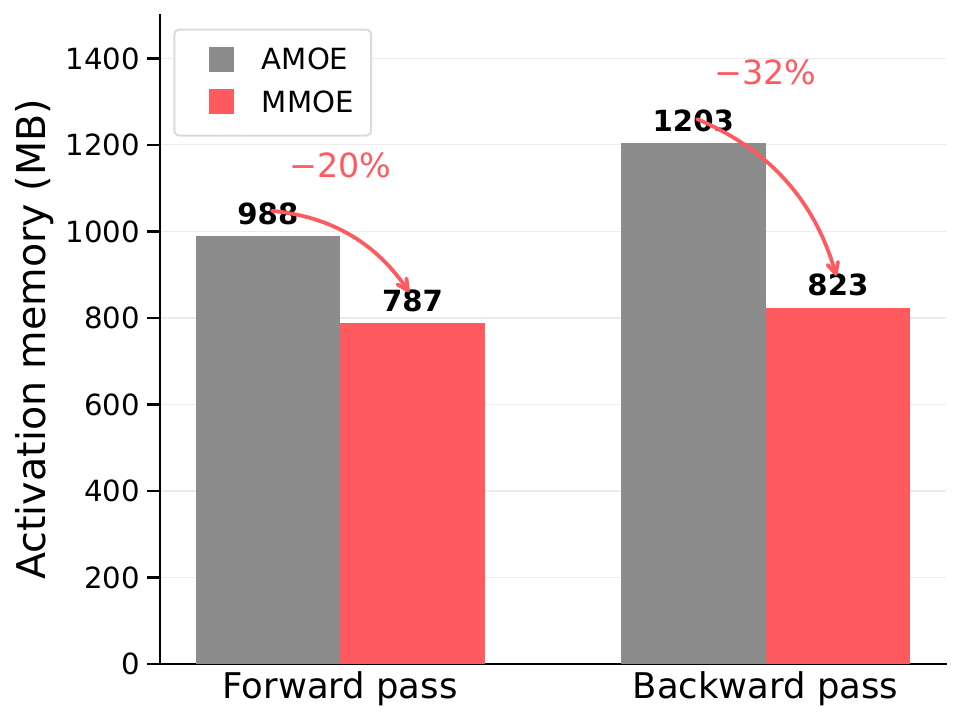}
    \caption{Per-block activation memory (MB) for an XL/2 block: attention-residual MoE (AMOE) versus the MMOE lightweight-expert design at matched 42.5M parameters. Routing some tokens to copy, zero, or constant experts cuts forward memory by about 20\% and backward memory by about 32\%.}
    \label{fig:memory-profile}
\end{figure}

The goal of MMOE is not only to reduce FID, but also to rebalance model capacity, activated computation, memory footprint, and training time. The training-time row in Table~\ref{tab:main-imagenet256} shows that architectural choices have very different system costs. Standard MoE increases training time substantially relative to dense SiT because routing and expert dispatch introduce overhead. AMOE achieves strong quality but is the most expensive measured variant. MMOE retains the quality benefit of attention-residual information reuse while reducing the cost by replacing some heavy expert routes with lightweight expert functions.

Figure~\ref{fig:memory-profile} provides a block-level memory comparison between AMOE and the MMOE-style lightweight expert design. Both configurations use an XL/2 block with hidden width 1152 and the same 42.5M parameters, but the lightweight expert design reduces single-block activation memory from 988~MB to 787~MB in the forward pass (about 20\%) and from 1203~MB to 823~MB in the backward pass (about 32\%). Because a substantial fraction of route slots select the copy, zero, or constant experts (44.6\% in early blocks, decreasing to 23.4\% in late blocks; Table~\ref{tab:routing-stats}), the corresponding routes skip the activation storage and gradient computation of a full MLP. Aggregated over the 28 transformer blocks, this amounts to approximately 5.5~GB of forward and over 10~GB of backward activation memory saved. This supports the design motivation behind copy, zero, and constant experts: a sparse model should not force every selected route to execute a full MLP when cheaper transformations are sufficient.

Beyond per-block memory, we profiled the multi-GPU training of the sparse expert models to locate where wall-clock time is spent. A single profiling run indicates that collective communication dominates: \texttt{ncclAllReduce} accounts for almost all of the profiled communication overhead, and device synchronization (\texttt{cudaStreamSynchronize}) accounts for roughly half of the total profiled time. The current training is therefore communication bound rather than compute bound, so the wall-clock figures in Table~\ref{tab:main-imagenet256} reflect a communication-limited regime. This also implies that part of the cost gap between dense and sparse variants stems from expert dispatch and gradient synchronization rather than from raw expert computation, and that a more efficient distributed implementation is a concrete avenue for reducing the training cost of sparse AFMs.

\begin{table}[tb]
    \centering
    \caption{FID sensitivity to the number of generated evaluation samples for MMOE-XL/2 (400k steps, 50-step ODE). The estimate drops sharply from 10k to 50k samples, so low-sample FID should be read as a diagnostic rather than a headline number.}
    \label{tab:sample-count}
    \renewcommand{\arraystretch}{1.05}
    \begin{tabular}{l|ccccc}
        \toprule
        Samples & 10k & 20k & 30k & 40k & 50k \\
        \midrule
        FID & 6.64 & 4.85 & 4.04 & 3.82 & 3.75 \\
        \bottomrule
    \end{tabular}
\end{table}

Table~\ref{tab:sample-count} studies the effect of the number of generated samples used for FID estimation. The 10k-sample estimate is directionally consistent with the 50k-sample estimate for the checked checkpoint, but the large gap between the 10k estimate (6.64) and the 50k estimate (3.75) shows that low-sample FID should be treated as a diagnostic rather than a headline metric. Therefore, the main comparison in Table~\ref{tab:main-imagenet256} keeps the evaluation protocol fixed across models, and 50k-sample FID remains the preferred setting whenever compute allows.

\begin{table}[tb]
    \centering
    \caption{Sampling-step sensitivity for MMOE-XL/2 (400k steps, 10k samples). Increasing ODE steps beyond 50 does not improve FID for this checkpoint; the SDE column is a reference point at matched step counts.}
    \label{tab:sampling-steps}
    \renewcommand{\arraystretch}{1.05}
    \begin{tabular}{c|cc}
        \toprule
        Inference Steps & ODE FID & SDE FID \\
        \midrule
        50 & 6.64 & 6.49 \\
        100 & 6.66 & 6.33 \\
        150 & 6.63 & 6.48 \\
        200 & 6.62 & 6.41 \\
        250 & 6.63 & 6.58 \\
        \bottomrule
    \end{tabular}
\end{table}

Table~\ref{tab:sampling-steps} evaluates the number of inference steps for the MMOE-XL/2 checkpoint. Increasing ODE steps beyond 50 does not improve the result under the current configuration (ODE FID stays within 6.62--6.66), and the SDE sampler reaches only slightly lower FID than the 50-step ODE setting (6.33--6.58 versus 6.64) rather than a large gain. The absence of improvement with more ODE steps is mildly counterintuitive, because additional integration steps should reduce discretization error; this flat trend more likely reflects a confound with the guidance scale or the 10k-sample FID estimate used in this sweep than a genuine property of the sampler, and it warrants a controlled re-run at a fixed guidance scale and 50k samples. We therefore keep 50-step ODE sampling as the default protocol for the main comparison for consistency, and treat this table as a sampling-protocol observation for the measured checkpoint rather than a general claim about all samplers.

\begin{table}[tb]
    \centering
    \caption{Classifier-free guidance sensitivity (250-step ODE, 400k steps). Lowering the guidance scale from 1.5 to 1.0 degrades all models, but MMOE remains the strongest at both settings.}
    \label{tab:cfg}
    \renewcommand{\arraystretch}{1.05}
    \begin{tabular}{c|cc>{\columncolor{tvcgblue!60}}c}
        \toprule
        CFG scale & SiT & SMOE & MMOE \\
        \midrule
        1.5 & 4.91 & 4.39 & 3.60 \\
        1.0 & 18.90 & 17.00 & 13.40 \\
        \bottomrule
    \end{tabular}
\end{table}

Table~\ref{tab:cfg} evaluates classifier-free guidance sensitivity. Reducing the guidance scale from 1.5 to 1.0 degrades all compared models under this 250-step setting, showing that absolute FID remains sensitive to guidance and sampler choices. Nevertheless, MMOE remains the strongest among the compared models at both guidance scales, suggesting that the architectural improvement is not tied to a single guidance setting.

\subsection{Visualization}
\label{subsec:visualization}

\begin{figure*}[t]
    \centering
    \includegraphics[width=\linewidth]{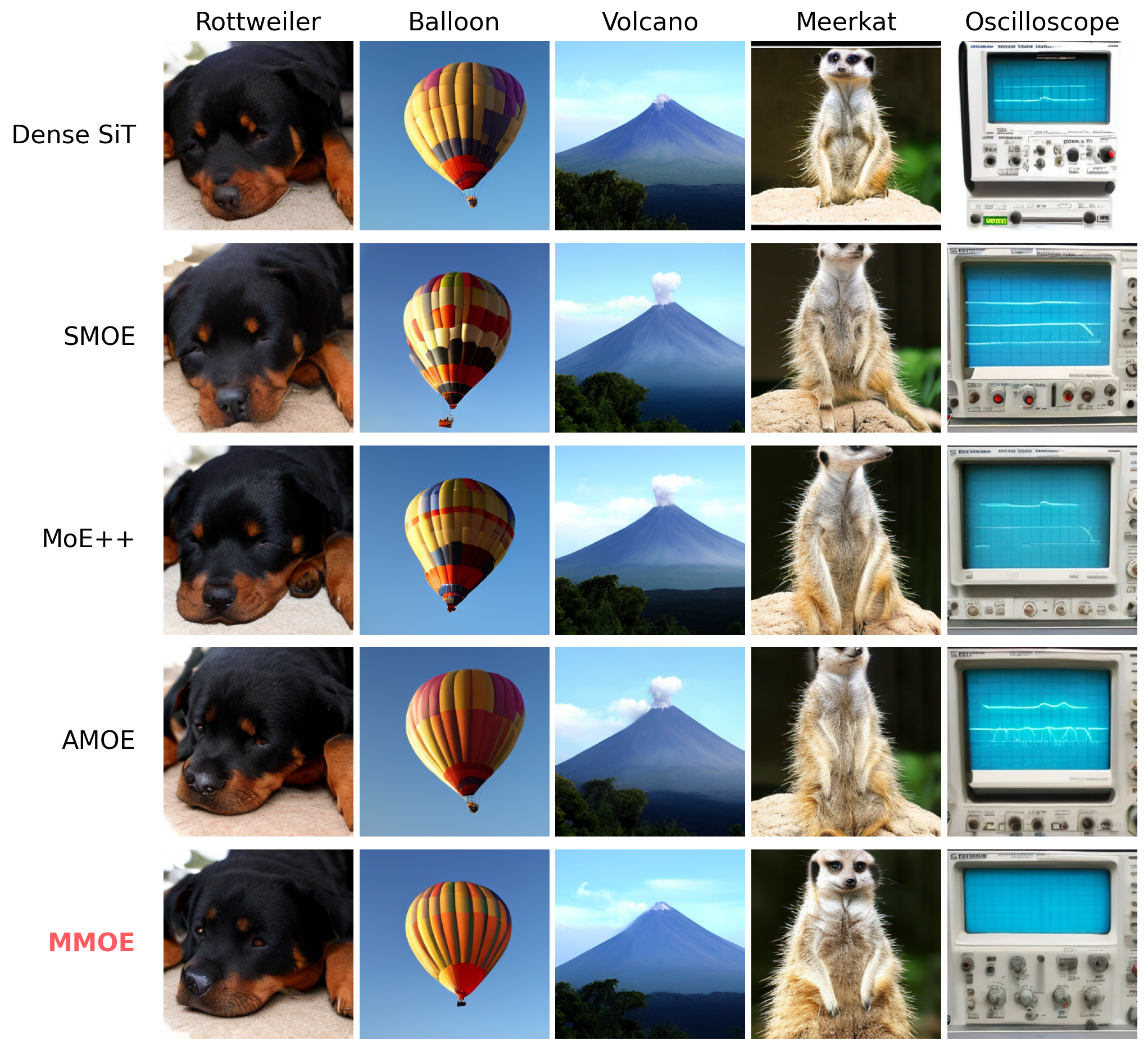}
    \caption{Qualitative comparison along the modernization path (top to bottom: Dense SiT, SMOE, MoE++, AMOE, MMOE) on five ImageNet classes. All rows share the same initial noise and class label and use 250-step ODE sampling at guidance scale 4.0, so differences reflect the architecture rather than the sample.}
    \label{fig:qualitative}
\end{figure*}

Figure~\ref{fig:qualitative} provides a qualitative comparison in which the five variants use the same initial noise and class labels, 250-step ODE sampling, and classifier-free guidance scale 4.0, with samples decoded from VAE latents by the shared decoder; this guidance scale is chosen for visualization and differs from the scales used in the quantitative tables. The qualitative samples are intended to complement FID rather than replace it: they help inspect whether the quantitative gains along the modernization path correspond to visible improvements in image structure and fidelity, while the controlled tables above provide the primary evidence for architectural comparison.

\section{Expert Routing Analysis}
\label{sec:routing-analysis}

We analyze MMOE routing with forward hooks on the expert routers during denoising. The analysis records selected experts for each step, block, image, token, and top-$k$ route. Table~\ref{tab:routing-stats} reports statistics for the converged MMOE-XL/2 checkpoint at 400k training steps, measured over 10k randomly sampled class-conditional denoising trajectories with 50 ODE steps and no classifier-free guidance. The expert pool contains four standard MLP experts and four lightweight experts: two constant experts, one copy expert, and one zero expert. In the table, routing entropy is computed from hard-dispatch expert loads and normalized by $\log E$ with $E=8$; the lightweight route fraction counts dispatch slots assigned to the two constant, copy, or zero experts; and the switch rate is the fraction of tokens whose top-$2$ expert set changes between adjacent denoising steps.

\paragraph{Denoising-time routing stability.}
The measured step-to-step switch rate is low. The top-$k$ expert set changes for only 1.43\%, 1.78\%, and 2.71\% of tokens in early, middle, and late block groups, respectively. Thus, the learned routing is not highly volatile across adjacent denoising steps. Instead, the model tends to keep stable expert sets along the denoising trajectory, with slightly more temporal adaptation in later blocks. This behavior is consistent with gate-residual routing: routing decisions can remain coherent across depth and denoising time while still allowing local changes when the token representation evolves.

\paragraph{Aggregate expert utilization.}
The load statistics indicate strong block-wise specialization rather than uniform expert usage. Across block groups, the load Gini coefficient is about 0.71--0.72 and the normalized load entropy is about 0.43--0.45. Under top-2 routing with eight experts, this corresponds to an effective usage of roughly 2.5 experts per block on average. This is not a single-expert collapse, because the dominant expert typically receives at most about half of all dispatches, but it does show that each block learns a small preferred expert subset. The lightweight route fraction also changes with depth: lightweight experts account for 44.6\% of route slots in early blocks, 34.8\% in middle blocks, and 23.4\% in late blocks. This depth trend supports the intended role of lightweight experts: early computation often selects cheap copy, zero, or constant transformations, while later blocks rely more heavily on standard MLP experts for refinement.

\begin{table*}[t]
    \centering
    \caption{Quantitative routing statistics for the converged MMOE-XL/2 model, measured over 10k denoising trajectories (50 ODE steps, no guidance). The high load Gini with moderate entropy indicates block-wise expert specialization, and the lightweight route fraction decreases with depth.}
    \label{tab:routing-stats}
    \renewcommand{\arraystretch}{1.15}
    \begin{tabular}{c|cccc}
        \toprule
        Block group & Load Gini & Routing Entropy & Lightweight Route Frac. & Switch Rate \\
        \midrule
        Early (0--8)   & $0.7206 \pm 0.0450$ & $0.4324 \pm 0.1118$ & 0.4457 & 0.0143 \\
        Middle (9--18) & $0.7198 \pm 0.0302$ & $0.4390 \pm 0.0789$ & 0.3482 & 0.0178 \\
        Late (19--27)  & $0.7115 \pm 0.0414$ & $0.4532 \pm 0.0839$ & 0.2341 & 0.0271 \\
        \bottomrule
    \end{tabular}
\end{table*}

\section{Discussion and Limitations}
\label{sec:discussion}

MMOE improves the quality-efficiency tradeoff of diffusion transformers by combining sparse routing with lightweight expert functions and attention-residual information reuse. The empirical gains are strongest on ImageNet-256, where MMOE consistently outperforms the dense SiT baseline and the intermediate sparse-expert variants we study under a matched evaluation protocol. We re-implement all compared variants ourselves, so these are controlled internal comparisons rather than a benchmark against published systems. The controlled modernization path also clarifies which components matter: sparse expert capacity helps, but the larger gain appears when lightweight routes and cross-depth aggregation are introduced. More broadly, by importing efficiency mechanisms proven in LLM MoE rather than enlarging the total expert pool, MMOE improves generation quality over the dense and intermediate sparse-expert baselines it studies within a single-node eight-GPU H100 budget (batch size 256, 400k steps), which suggests that AFMs can follow the balanced scaling path taken by LLMs instead of inflating total parameters and sparsity ratios.

The current evidence has several limitations. First, the main experiments are class-conditional ImageNet experiments, so the results do not establish general text-to-image superiority. Second, all reported comparisons are internal: we re-implement the dense and intermediate sparse-expert baselines under a matched recipe, and a head-to-head comparison against published diffusion-MoE methods (Table~\ref{tab:external-baselines}) is still in progress. Third, most FID values are single-run estimates; the seed-robustness study in Table~\ref{tab:seed-robustness} shows that for the XL/2 path the between-variant gaps exceed the seed-induced standard deviation, but per-seed variance for the other scales and studies is not measured, so small gaps elsewhere should still be read with caution. Fourth, the resolution study covers ImageNet-256 and, for the dense-versus-MMOE comparison, ImageNet-512 (Table~\ref{tab:imagenet512}), but the full modernization path and the model-scale study are established at 256, so conclusions at higher resolutions are limited. Fifth, the routing analysis is measured on the converged XL checkpoint under the 50-step no-guidance analysis protocol; extending it to other model sizes, guidance scales, and training stages would further clarify how stable the observed specialization pattern is. Finally, sparse expert models introduce distributed-training overheads that block-level memory profiling and single-run wall-clock time do not fully capture; our profiling indicates that the current multi-GPU training is communication bound, so the reported training times reflect a communication-limited regime.

Future work includes extending MMOE to large-scale text-to-image training, studying routing under multimodal conditioning, and improving distributed training efficiency for sparse expert diffusion models.

\section{Conclusion}
\label{sec:conclusion}

We presented MMOE, a controlled modernization of SiT-style diffusion transformers for AIGC Foundation Models. Rather than following current AFM scaling by inflating total parameters and sparsity ratios, MMOE imports efficiency mechanisms proven in LLM MoE, combining routed MLP experts, lightweight copy/zero/constant experts, gate-residual routing, and attention-residual aggregation. On class-conditional ImageNet-256 generation, and within a single-node eight-GPU H100 budget (batch size 256, 400k steps), the modernization path improves FID and convergence over dense SiT and the intermediate MoE variants under matched training and sampling protocols; among the sparse variants MMOE gives the best quality-cost balance, cutting the training time of the most expensive variant (AMOE) from 120h to 67h, though it remains more expensive than the dense baseline. The routing analysis further suggests that MMOE learns stable block-wise expert specialization, uses lightweight routes substantially in early and middle blocks, and changes its top-$2$ routing set only modestly between adjacent denoising steps. These findings support sparse expert modernization as a practical direction for improving the quality-cost balance of diffusion transformers while preserving the standard training and sampling interface.

\section*{Acknowledgments}
This research is supported by the Institute of Artificial Intelligence of China Telecom (TeleAI). This research is supported by the RIE2025 Industry Alignment Fund – Industry Collaboration Projects (IAF-ICP) (Award I2301E0026), administered by A*STAR, as well as supported by Alibaba Group and NTU Singapore through Alibaba-NTU Global e-Sustainability CorpLab (ANGEL). The work is also supported by the Ministry of Education, Singapore under its MOE Academic Research Fund Tier 2 (MOE-T2EP20123-0005).


\newpage
{
    \clearpage
    \bibliographystyle{IEEEtran}
    \bibliography{reference}

@inproceedings{liu2022convnet,
  title = {A ConvNet for the 2020s},
  author = {Liu, Zhuang and Mao, Hanzi and Wu, Chao-Yuan and Feichtenhofer, Christoph and Darrell, Trevor and Xie, Saining},
  booktitle = {Proceedings of the IEEE/CVF Conference on Computer Vision and Pattern Recognition (CVPR)},
  pages = {11976--11986},
  year = {2022}
}

@inproceedings{deng2009imagenet,
  title = {{ImageNet}: A Large-Scale Hierarchical Image Database},
  author = {Deng, Jia and Dong, Wei and Socher, Richard and Li, Li-Jia and Li, Kai and Fei-Fei, Li},
  booktitle = {Proceedings of the IEEE Conference on Computer Vision and Pattern Recognition (CVPR)},
  pages = {248--255},
  year = {2009},
  doi = {10.1109/CVPR.2009.5206848}
}

@inproceedings{heusel2017gans,
  title = {{GANs} Trained by a Two Time-Scale Update Rule Converge to a Local Nash Equilibrium},
  author = {Heusel, Martin and Ramsauer, Hubert and Unterthiner, Thomas and Nessler, Bernhard and Hochreiter, Sepp},
  booktitle = {Advances in Neural Information Processing Systems (NeurIPS)},
  year = {2017}
}

@inproceedings{rombach2022high,
  title = {High-Resolution Image Synthesis with Latent Diffusion Models},
  author = {Rombach, Robin and Blattmann, Andreas and Lorenz, Dominik and Esser, Patrick and Ommer, Bj{\"o}rn},
  booktitle = {Proceedings of the IEEE/CVF Conference on Computer Vision and Pattern Recognition (CVPR)},
  pages = {10684--10695},
  year = {2022}
}

@inproceedings{ho2020denoising,
  title = {Denoising Diffusion Probabilistic Models},
  author = {Ho, Jonathan and Jain, Ajay and Abbeel, Pieter},
  booktitle = {Advances in Neural Information Processing Systems (NeurIPS)},
  volume = {33},
  pages = {6840--6851},
  year = {2020}
}

@inproceedings{song2021score,
  title = {Score-Based Generative Modeling through Stochastic Differential Equations},
  author = {Song, Yang and Sohl-Dickstein, Jascha and Kingma, Diederik P. and Kumar, Abhishek and Ermon, Stefano and Poole, Ben},
  booktitle = {International Conference on Learning Representations (ICLR)},
  year = {2021}
}

@inproceedings{lipman2023flow,
  title = {Flow Matching for Generative Modeling},
  author = {Lipman, Yaron and Chen, Ricky T. Q. and Ben-Hamu, Heli and Nickel, Maximilian and Le, Matt},
  booktitle = {International Conference on Learning Representations (ICLR)},
  year = {2023}
}

@article{albergo2025stochastic,
  title = {Stochastic Interpolants: A Unifying Framework for Flows and Diffusions},
  author = {Albergo, Michael S. and Boffi, Nicholas M. and Vanden-Eijnden, Eric},
  journal = {Journal of Machine Learning Research},
  volume = {26},
  number = {209},
  pages = {1--80},
  year = {2025}
}

@inproceedings{peebles2023scalable,
  title = {Scalable Diffusion Models with Transformers},
  author = {Peebles, William and Xie, Saining},
  booktitle = {Proceedings of the IEEE/CVF International Conference on Computer Vision (ICCV)},
  pages = {4172--4182},
  year = {2023}
}

@inproceedings{ma2024sit,
  title = {{SiT}: Exploring Flow and Diffusion-Based Generative Models with Scalable Interpolant Transformers},
  author = {Ma, Nanye and Goldstein, Mark and Albergo, Michael S. and Boffi, Nicholas M. and Vanden-Eijnden, Eric and Xie, Saining},
  booktitle = {European Conference on Computer Vision (ECCV)},
  year = {2024}
}

@misc{ho2022classifierfree,
  title = {Classifier-Free Diffusion Guidance},
  author = {Ho, Jonathan and Salimans, Tim},
  year = {2022},
  eprint = {2207.12598},
  archivePrefix = {arXiv},
  primaryClass = {cs.LG}
}

@inproceedings{shazeer2017outrageously,
  title = {Outrageously Large Neural Networks: The Sparsely-Gated Mixture-of-Experts Layer},
  author = {Shazeer, Noam and Mirhoseini, Azalia and Maziarz, Krzysztof and Davis, Andy and Le, Quoc and Hinton, Geoffrey and Dean, Jeff},
  booktitle = {International Conference on Learning Representations (ICLR)},
  year = {2017}
}

@inproceedings{lepikhin2021gshard,
  title = {{GShard}: Scaling Giant Models with Conditional Computation and Automatic Sharding},
  author = {Lepikhin, Dmitry and Lee, HyoukJoong and Xu, Yuanzhong and Chen, Dehao and Firat, Orhan and Huang, Yanping and Krikun, Maxim and Shazeer, Noam and Chen, Zhifeng},
  booktitle = {International Conference on Learning Representations (ICLR)},
  year = {2021}
}

@article{fedus2022switch,
  title = {Switch Transformers: Scaling to Trillion Parameter Models with Simple and Efficient Sparsity},
  author = {Fedus, William and Zoph, Barret and Shazeer, Noam},
  journal = {Journal of Machine Learning Research},
  volume = {23},
  number = {120},
  pages = {1--39},
  year = {2022}
}

@misc{jiang2024mixtral,
  title = {Mixtral of Experts},
  author = {Jiang, Albert Q. and Sablayrolles, Alexandre and Roux, Antoine and Mensch, Arthur and Savary, Blanche and Bamford, Chris and Singh Chaplot, Devendra and de las Casas, Diego and others},
  year = {2024},
  eprint = {2401.04088},
  archivePrefix = {arXiv},
  primaryClass = {cs.LG}
}

@inproceedings{jin2025moepp,
  title = {{MoE++}: Accelerating Mixture-of-Experts Methods with Zero-Computation Experts},
  author = {Jin, Peng and Zhu, Bo and Yuan, Li and Yan, Shuicheng},
  booktitle = {International Conference on Learning Representations (ICLR)},
  year = {2025}
}

@misc{kimiteam2026attentionresiduals,
  title = {Attention Residuals},
  author = {{Kimi Team} and Chen, Guangyu and Zhang, Yu and Su, Jianlin and Xu, Weixin and Pan, Siyuan and others},
  year = {2026},
  eprint = {2603.15031},
  archivePrefix = {arXiv},
  primaryClass = {cs.CL}
}

@misc{fei2024scaling,
  title = {Scaling Diffusion Transformers to 16 Billion Parameters},
  author = {Fei, Zhengcong and Fan, Mingyuan and Yu, Changqian and Li, Debang and Huang, Junshi},
  year = {2024},
  eprint = {2407.11633},
  archivePrefix = {arXiv},
  primaryClass = {cs.CV}
}

@inproceedings{sun2025ecdit,
  title = {{EC-DIT}: Scaling Diffusion Transformers with Adaptive Expert-Choice Routing},
  author = {Sun, Haotian and Lei, Tao and Zhang, Bowen and Li, Yanghao and Huang, Haoshuo and Pang, Ruoming and Dai, Bo and Du, Nan},
  booktitle = {International Conference on Learning Representations (ICLR)},
  year = {2025}
}

@inproceedings{cheng2025diffmoe,
  title = {{Diff-MoE}: Diffusion Transformer with Time-Aware and Space-Adaptive Experts},
  author = {Cheng, Kun and He, Xiao and Yu, Lei and Tu, Zhijun and Zhu, Mingrui and Wang, Nannan and Gao, Xinbo and Hu, Jie},
  booktitle = {Proceedings of the 42nd International Conference on Machine Learning},
  series = {Proceedings of Machine Learning Research},
  volume = {267},
  pages = {10010--10024},
  year = {2025}
}

@misc{yuan2025race,
  title = {Expert Race: A Flexible Routing Strategy for Scaling Diffusion Transformer with Mixture of Experts},
  author = {Yuan, Yike and Wang, Ziyu and Huang, Zihao and Zhu, Defa and Zhou, Xun and Yu, Jingyi and Min, Qiyang},
  year = {2025},
  eprint = {2503.16057},
  archivePrefix = {arXiv},
  primaryClass = {cs.CV}
}

@misc{wei2025routing,
  title = {Routing Matters in {MoE}: Scaling Diffusion Transformers with Explicit Routing Guidance},
  author = {Wei, Yujie and Zhang, Shiwei and Yuan, Hangjie and Han, Yujin and Chen, Zhekai and Wang, Jiayu and Zou, Difan and Liu, Xihui and Zhang, Yingya and Liu, Yu and Shan, Hongming},
  year = {2025},
  eprint = {2510.24711},
  archivePrefix = {arXiv},
  primaryClass = {cs.CV}
}

@misc{liu2025efficientmoe,
  title = {Efficient Training of Diffusion Mixture-of-Experts Models: A Practical Recipe},
  author = {Liu, Yahui and Yue, Yang and Zhang, Jingyuan and Sun, Chenxi and Zhou, Yang and Zeng, Wencong and Tang, Ruiming and Zhou, Guorui},
  year = {2025},
  eprint = {2512.01252},
  archivePrefix = {arXiv},
  primaryClass = {cs.LG}
}

@misc{shi2025diffmoe,
  title = {{DiffMoE}: Dynamic Token Selection for Scalable Diffusion Transformers},
  author = {Shi, Minglei and Yuan, Ziyang and Yang, Haotian and Wang, Xintao and Zheng, Mingwu and Tao, Xin and Zhao, Wenliang and Zheng, Wenzhao and Zhou, Jie and Lu, Jiwen and Wan, Pengfei and Zhang, Di and Gai, Kun},
  year = {2025},
  eprint = {2503.14487},
  archivePrefix = {arXiv},
  primaryClass = {cs.CV}
}

@inproceedings{li2026back,
  title={Back to basics: Let denoising generative models denoise},
  author={Li, Tianhong and He, Kaiming},
  booktitle={Proceedings of the IEEE/CVF Conference on Computer Vision and Pattern Recognition},
  pages={36115--36125},
  year={2026}
}

@inproceedings{zhao2025dynamic,
  title={Dynamic diffusion transformer},
  author={Zhao, Wangbo and Han, Yizeng and Tang, Jiasheng and Wang, Kai and Song, Yibing and Huang, Gao and Wang, Fan and You, Yang},
  booktitle={International Conference on Learning Representations},
  volume={2025},
  pages={65520--65552},
  year={2025}
}

@article{shao2025ai,
  title={Ai flow at the network edge},
  author={Shao, Jiawei and Li, Xuelong},
  journal={IEEE Network},
  year={2025},
  publisher={IEEE}
}

@article{an2026ai,
  title={Ai flow: Perspectives, scenarios, and approaches},
  author={An, Hongjun and Hu, Wenhan and Huang, Sida and Huang, Siqi and Li, Ruanjun and Liang, Yuanzhi and Shao, Jiawei and Song, Yiliang and Wang, Zihan and Yuan, Cheng and others},
  journal={Vicinagearth},
  volume={3},
  number={1},
  pages={1},
  year={2026},
  publisher={Springer}
}

@misc{chen2026generativetransmissionrethinkingcomputation,
      title={Generative Transmission: Rethinking Computation, Bandwidth, and Memory in Communication}, 
      author={Xiangyu Chen and Jixiang Luo and Yuankai Fan and Haibin Huang and Chi Zhang and Xuelong Li},
      year={2026},
      eprint={2607.17482},
      archivePrefix={arXiv},
      primaryClass={cs.CV}, 
}

@article{liu2023nero,
  title={Nero: Neural geometry and brdf reconstruction of reflective objects from multiview images},
  author={Liu, Yuan and Wang, Peng and Lin, Cheng and Long, Xiaoxiao and Wang, Jiepeng and Liu, Lingjie and Komura, Taku and Wang, Wenping},
  journal={ACM Transactions on Graphics (ToG)},
  volume={42},
  number={4},
  pages={1--22},
  year={2023},
  publisher={ACM New York, NY, USA}
}

@article{wang2025mmgen,
  title={Mmgen: Unified multi-modal image generation and understanding in one go},
  author={Wang, Jiepeng and Wang, Zhaoqing and Pan, Hao and Liu, Yuan and Yu, Dongdong and Wang, Changhu and Wang, Wenping},
  journal={arXiv preprint arXiv:2503.20644},
  year={2025}
}

@article{jia2026seeing,
  title={Seeing sound, hearing sight: Uncovering modality bias and conflict of ai models in sound localization},
  author={Jia, Yanhao and Xie, Ji and Jivaganesh, S and Hao, Li and Wu, Xu and Zhang, Mengmi},
  journal={Advances in neural information processing systems},
  volume={38},
  pages={148468--148499},
  year={2026}
}

@inproceedings{jia2025uni,
  title={Uni-retrieval: A multi-style retrieval framework for stem’s education},
  author={Jia, Yanhao and Wu, Xinyi and Hao, Li and Qinglin, Zhang and Hu, Yuxiao and Zhao, Shuai and Fan, Wenqi},
  booktitle={Proceedings of the 63rd Annual Meeting of the Association for Computational Linguistics (Volume 1: Long Papers)},
  pages={10182--10197},
  year={2025}
}

@article{wu2026towards,
  title={Towards Affective Evaluation of STEM Education: Leveraging MLLMs in Project-Based Learning},
  author={Wu, Xinyi and Jia, Yanhao and Zhang, Qinglin and Qin, Yiran and Xiao, Luwei and Zhao, Shuai},
  journal={IEEE Transactions on Affective Computing},
  year={2026},
  publisher={IEEE}
}

@article{wu2025query,
  title={From query to explanation: Uni-rag for multi-modal retrieval-augmented learning in stem},
  author={Wu, Xinyi and Jia, Yanhao and Xiao, Luwei and Zhao, Shuai and Chiang, Fengkuang and Cambria, Erik},
  journal={arXiv preprint arXiv:2507.03868},
  year={2025}
}

@inproceedings{xi2026omnivdiff,
  title={Omnivdiff: Omni controllable video diffusion for generation and understanding},
  author={Xi, Dianbing and Wang, Jiepeng and Liang, Yuanzhi and Qiu, Xi and Huo, Yuchi and Wang, Rui and Zhang, Chi and Li, Xuelong},
  booktitle={Proceedings of the AAAI Conference on Artificial Intelligence},
  volume={40},
  number={13},
  pages={10915--10923},
  year={2026}
}

@article{xi2025ctrlvdiff,
  title={CtrlVDiff: Controllable Video Generation via Unified Multimodal Video Diffusion},
  author={Xi, Dianbing and Wang, Jiepeng and Liang, Yuanzhi and Qiu, Xi and Liu, Jialun and Pan, Hao and Huo, Yuchi and Wang, Rui and Huang, Haibin and Zhang, Chi and others},
  journal={arXiv preprint arXiv:2511.21129},
  year={2025}
}

@article{wu2026edustory,
  title={EduStory: A Unified Framework for Pedagogically-Consistent Multi-Shot STEM Instructional Video Generation},
  author={Wu, Xinyi and Teotia, Jayant and Zhao, Shuai and Cambria, Erik},
  journal={arXiv preprint arXiv:2605.09378},
  year={2026}
}

@inproceedings{du2022glam,
  title = {{GLaM}: Efficient Scaling of Language Models with Mixture-of-Experts},
  author = {Du, Nan and Huang, Yanping and Dai, Andrew M. and Tong, Simon and Lepikhin, Dmitry and Xu, Yuanzhong and Krikun, Maxim and Zhou, Yanqi and Yu, Adams Wei and Firat, Orhan and Zoph, Barret and Fedus, Liam and Bosma, Maarten and Zhou, Zongwei and Wang, Tao and Wang, Yu Emma and Webster, Kellie and Pellat, Marie and Robinson, Kevin and Meier-Hellstern, Kathleen and Duke, Toju and Dixon, Lucas and Zhang, Kun and Le, Quoc V. and Wu, Yonghui and Chen, Zhifeng and Cui, Claire},
  booktitle = {Proceedings of the 39th International Conference on Machine Learning (ICML)},
  series = {Proceedings of Machine Learning Research},
  volume = {162},
  pages = {5547--5569},
  year = {2022}
}

@misc{zoph2022stmoe,
  title = {{ST-MoE}: Designing Stable and Transferable Sparse Expert Models},
  author = {Zoph, Barret and Bello, Irwan and Kumar, Sameer and Du, Nan and Huang, Yanping and Dean, Jeff and Shazeer, Noam and Fedus, William},
  year = {2022},
  eprint = {2202.08906},
  archivePrefix = {arXiv},
  primaryClass = {cs.LG}
}

@inproceedings{dai2024deepseekmoe,
  title = {{DeepSeekMoE}: Towards Ultimate Expert Specialization in Mixture-of-Experts Language Models},
  author = {Dai, Damai and Deng, Chengqi and Zhao, Chenggang and Xu, R. X. and Gao, Huazuo and Chen, Deli and Li, Jiashi and Zeng, Wangding and Yu, Xingkai and Wu, Y. and Xie, Zhenda and Li, Y. K. and Huang, Panpan and Luo, Fuli and Ruan, Chong and Sui, Zhifang and Liang, Wenfeng},
  booktitle = {Proceedings of the 62nd Annual Meeting of the Association for Computational Linguistics (ACL)},
  pages = {1280--1297},
  year = {2024}
}

@inproceedings{zhou2022mixture,
  title = {Mixture-of-Experts with Expert Choice Routing},
  author = {Zhou, Yanqi and Lei, Tao and Liu, Hanxiao and Du, Nan and Huang, Yanping and Zhao, Vincent and Dai, Andrew M. and Chen, Zhifeng and Le, Quoc V. and Laudon, James},
  booktitle = {Advances in Neural Information Processing Systems (NeurIPS)},
  year = {2022}
}

@misc{deepseekai2024deepseekv3,
  title = {{DeepSeek-V3} Technical Report},
  author = {{DeepSeek-AI}},
  year = {2024},
  eprint = {2412.19437},
  archivePrefix = {arXiv},
  primaryClass = {cs.CL}
}

@misc{raposo2024mixture,
  title = {Mixture-of-Depths: Dynamically Allocating Compute in Transformer-Based Language Models},
  author = {Raposo, David and Ritter, Sam and Richards, Blake and Lillicrap, Timothy and Humphreys, Peter Conway and Santoro, Adam},
  year = {2024},
  eprint = {2404.02258},
  archivePrefix = {arXiv},
  primaryClass = {cs.LG}
}

@inproceedings{riquelme2021scaling,
  title = {Scaling Vision with Sparse Mixture of Experts},
  author = {Riquelme, Carlos and Puigcerver, Joan and Mustafa, Basil and Neumann, Maxim and Jenatton, Rodolphe and Susano Pinto, Andr\'{e} and Keysers, Daniel and Houlsby, Neil},
  booktitle = {Advances in Neural Information Processing Systems (NeurIPS)},
  year = {2021}
}

@inproceedings{bao2023all,
  title = {All are Worth Words: A {ViT} Backbone for Diffusion Models},
  author = {Bao, Fan and Nie, Shen and Xue, Kaiwen and Cao, Yue and Li, Chongxuan and Su, Hang and Zhu, Jun},
  booktitle = {Proceedings of the IEEE/CVF Conference on Computer Vision and Pattern Recognition (CVPR)},
  year = {2023}
}

@inproceedings{karras2022elucidating,
  title = {Elucidating the Design Space of Diffusion-Based Generative Models},
  author = {Karras, Tero and Aittala, Miika and Aila, Timo and Laine, Samuli},
  booktitle = {Advances in Neural Information Processing Systems (NeurIPS)},
  year = {2022}
}

@inproceedings{yu2025representation,
  title = {Representation Alignment for Generation: Training Diffusion Transformers Is Easier Than You Think},
  author = {Yu, Sihyun and Kwak, Sangkyung and Jang, Huiwon and Jeong, Jongheon and Huang, Jonathan and Shin, Jinwoo and Xie, Saining},
  booktitle = {International Conference on Learning Representations (ICLR)},
  year = {2025}
}

@misc{jiatowards,
  title={Towards Spatial Reasoning and Understanding via Modeling Modality Conflict, Bias and Alignment},
  author={Jia, Yanhao and Wu, Xinyi and Teotia, Jayant and Dong, Junhao and Zhao, Shuai and Koniusz, Piotr and Cambria, Erik},
  year = {2026}
}

@article{cambria2026senticnet,
  title={SenticNet 9: Generative commonsense for emotion AI via conceptual primitive discovery and time shift mechanism},
  author={Cambria, Erik and Mao, Rui and Zhang, Xulang and Xiao, Luwei and Shen, Tiesunlong and Anand, Avinash},
  journal={IEEE Transactions on Computational Social Systems},
  year={2026},
  publisher={IEEE}
}

@article{kukreja2026forge,
  title={FORGE: Fused On-Register Gradient Elimination for Memory-Efficient LLM Training},
  author={Kukreja, Dikshant and Prasad, Kritarth and Anand, Avinash and Wang, Zhengkui and Cambria, Erik and Liu, Timothy and Ng, Aik Beng and See, Simon and Chatterjee, Bapi},
  journal={arXiv preprint arXiv:2606.22932},
  year={2026}
}
}

\end{document}